%% file: root.tex
\documentclass[letterpaper, 10 pt, conference]{ieeeconf}  % Comment this line out if you need a4paper

\IEEEoverridecommandlockouts                              % This command is only needed if 
\usepackage{times}

\usepackage{algorithm}
\usepackage[noend]{algcompatible}

\usepackage{amsmath}
\usepackage{amsfonts}
\usepackage{breqn}
\usepackage{multirow,hhline,graphicx,array}
\graphicspath{ {./images/} }
\usepackage{multicol}
\usepackage[bookmarks=true]{hyperref}
\usepackage{subfiles}
\usepackage{caption}
\usepackage{xcolor}
\usepackage{footnote}
\usepackage[flushleft]{threeparttable}

\title{\LARGE \bf
LiDARNet: A Boundary-Aware Domain Adaptation Model for Point Cloud Semantic Segmentation
}

\author{Peng Jiang$^{1}$ and Srikanth Saripalli$^{1}$% <-this % stops a space
\thanks{$^{1}$Peng Jiang ({\tt\small maskjp@tamu.edu}) and Srikanth Saripalli ({\tt\small  ssaripalli@tamu.edu}) are with the J. Mike Walker '66 Department of Mechanical Engineering, Texas A\&M University,
College Station, TX 77840, USA}%
}

\begin{document}

\maketitle
\thispagestyle{empty}
\pagestyle{empty}

%%%%%%%%%%%%%%%%%%%%%%%%%%%%%%%%%%%%%%%%%%%%%%%%%%%%%%%%%%%%%%%%%%%%%%%%%%%%%%%%
\input{sections/abstract.tex}
%%%%%%%%%%%%%%%%%%%%%%%%%%%%%%%%%%%%%%%%%%%%%%%%%%%%%%%%%%%%%%%%%%%%%%%%%%%%%%%%
\section{INTRODUCTION}
\subfile{sections/introduction}
\section{RELATED WORK}
\subfile{sections/related_work}
\section{OUR APPROACH}
\subfile{sections/our_approach}
\section{EXPERIMENTAL EVALUATION}
\subfile{sections/experimental_evaluation}
\section{CONCLUSIONS}
\subfile{sections/conclusion}
%\section*{APPENDIX}


%\section*{ACKNOWLEDGMENT}

%%%%%%%%%%%%%%%%%%%%%%%%%%%%%%%%%%%%%%%%%%%%%%%%%%%%%%%%%%%%%%%%%%%%%%%%%%%%%%%%

\bibliographystyle{IEEEtran}
\bibliography{IEEEabrv,references}

\end{document}

%% file: sections/abstract.tex
\begin{abstract}
     We present a boundary-aware domain adaptation model for LiDAR scan full-scene semantic segmentation (LiDARNet). Our model can extract both the domain private features and the domain shared features with a two branch structure.  We embedded Gated-SCNN into the segmentor component of LiDARNet to learn boundary information while learning to predict full-scene semantic segmentation labels. Moreover, we further reduce the domain gap by inducing the model to learn a mapping between two domains using the domain shared and private features. Additionally, we introduce a new dataset (SemanticUSL\footnote{The access address of SemanticUSL:\url{https://unmannedlab.github.io/research/SemanticUSL}}) for domain adaptation for LiDAR point cloud semantic segmentation. The dataset has the same data format and ontology as SemanticKITTI. We conducted experiments on real-world datasets SemanticKITTI, SemanticPOSS, and SemanticUSL, which have differences in channel distributions, reflectivity distributions, diversity of scenes, and sensors setup. Using our approach, we can get a single projection-based LiDAR full-scene semantic segmentation model working on both domains. Our model can keep almost the same performance on the source domain after adaptation and get an 8\%-22\% mIoU performance increase in the target domain. 
\end{abstract}

%% file: sections/introduction.tex
Scene understanding is important for autonomous robotics and vehicles. The semantic scene information can be used for navigation, decision making, and semantic mapping. Most autonomous vehicles these days are already equipped with cameras and LiDAR. The camera can provide dense color and texture information, making RGB images the first choice for semantic segmentation. However, cameras can be easily affected by varying illumination. Therefore semantic segmentation for LiDAR scans is also important.
The recent advancement of deep learning technology has lead to a significant development of semantic segmentation in 3D \cite{Guo2020}. Meanwhile, the release of several datasets \cite{Geyer2020,Che2019,Behley2019,Pan2020} from autonomous driving companies further promotes the research for LiDAR scan semantic segmentation. However, current learning-based methods mostly are supervised methods, of which training requires a large amount of annotated data. Besides, the performance of well-trained models can be hurt because of a slight departure of the test data from training data. For LiDAR scans, the departure can come from manufacture difference, sensor setup difference, scene contents difference.

Without annotating new data, using old labeled data to train a model that can work on new data is an unsupervised domain adaptation problem. The problem exists in both 2D image data and 3D point cloud data, but the characteristics of sparsity, irregularity and unstructured distribution  make this problem more difficult for the point cloud. To address the domain gap between two LiDAR scan dataset, we observe that LiDAR scans can be projected into 2D space and represented as range map. The projection reduces the difficulty of domain adaption caused by the sparsity, irregularity, and unstructured features of the LiDAR point cloud. Supervised learning algorithms based on this projection representation have achieved good performance on semantic segmentation for LiDAR scan\cite{Milioto2019,Cortinhal2020}. This observation motivates us chose a projection-based semantic segmentation model as the backbone and major task component of our model. 

Moreover, inspired by work on private-shared components separation \cite{Bousmalis2016}, we designed a model with two extractors: domain private extractor and domain shared extractor. The domain private extractor extracts private features of each domain, while the domain shared extractor extracts features sharing information across domains. In our case, the domain shared features contain the geometry and semantic meanings of objects and the content relationship of the scenes such as on-road scene. Meanwhile, the domain private features are the different noise distributions, point cloud distributions, and reflectivity distributions. To further reduce the domain gap, we not only train our models to use the shared features to perform our task: full scene semantic segmentation but also use the two separated features to learn two mappings between the source domain and target domain \(\{G: S\to T\}\) and \(\{G: T\to S\}\).  Furthermore, we notice that after the projection, the boundaries of objects are much easier to learn than in 3D space and can be beneficial for segmentation. 

Recent released datasets \cite{Geyer2020,Che2019,Behley2019,Pan2020} of LiDAR scan for autonomous driving are mostly collected on vehicles in traffic-road environments. To further verify our adaptation model's effectiveness, we introduce our LiDAR dataset  (SemanticUSL) for domain adaptation. Instead of using vehicles, our dataset was collected on a smaller robot with a 64-channels Ouster LiDAR. The data not only includes traffic-road scene but also have walkpath scenes and off-road scenes. We perform evaluation experiments with SemanticKITTI\cite{Behley2019}, SemanticPOSS, and our dataset. We also compared our results with the pixel-level CyCADA method \cite{Hoffman2018}.
The results show that our model has similar performance on the source domain after adaptation and has a 8\%-22\% improvement in mIoU in the target domain.

%% file: sections/related_work.tex
\subsection{Semantic segmentation}
Semantic segmentation is the task of assigning an object label to each basic unit of data like a pixel of an image or a point of a point cloud. Before the prevalence of deep learning, traditional segmentation methods of point cloud mainly relied on handcrafted features from geometric constraints and assumed prior knowledge \cite{Zhang2019_A_Review, Xie2019}. Due to the irregularity and lack of structure, point cloud has several representation forms, and it is not very easy to apply deep learning on point clouds directly. The simplest form is to represent point cloud as dense voxel grids and use 3D convolution to predict the results\cite{Zhou2018, Qi2016,Wu2015}. Dense voxel representation is redundant, and 3D convolution operation is also computationally inefficient. \cite{Retinskiy2019,Graham2018,Choy2019,Tang2020} have made an effort to reduce the computation cost and improve performance. Meanwhile, point cloud-based methods  \cite{Qi2017, Qi2017a,Thomas2019,Li2018} try to directly operate on point set to reduce computation cost and achieve good performance. Point clouds can also be represented as graphs and meshes further to explore the relationship between points.
Moreover, a LiDAR scan can also be projected on 2D space and represented as a dense range map without losing much information. The projection allows us to apply 2D convolutional directly on the LiDAR scan. The projection-based methods \cite{Milioto2019,Cortinhal2020,Wu2019} directly apply 2D convolution on the LiDAR scan and achieve good performance on accuracy. We chose the projection-based model as our domain adaptation model's task component and tried to improve its domain transferability. 
\subsection{Unsupervised Domain adaptation} 
Unsupervised Domain adaptation (UDA) is a subfield of transfer learning, to learn a discriminative model in the presence of domain shift between domains. Accompany with the development of deep learning, a lot of UDA method has been come up\cite{Long2015,Tzeng2014,Tzeng2017,Yu2020,Hoffman2018}. Hoffman et al. \cite{Hoffman2018} introduce the CycleGAN mechanism into the domain adaptation field and proposed a discriminatively-trained cycle consistent adversarial domain adaptation model (CyCADA).There are also some work specifically designed to address domain adaptation for point cloud. Wu et al. \cite{Wu2018seg} utilize geodesic correlation alignment to perform adaptation between real and synthetic LiDAR data. Rist et al.\cite{Rist2019} designed a voxel-based architecture to extract features of LiDAR point cloud and processed a supervised training methodology to learn traversable features. Salah et al. \cite{Saleh2019} convert LiDAR into bird-eye view images and used a CycleGAN method to adapt the synthetic lidar domain to real domains. Wang et al.\cite{Wang2019} also use the bird-eye view as representation and a two-scale model to perform cross-range adaption for LiDAR 3D object detection. Qin et al. \cite{Qin2019} propose a multi-scale 3D adaption network to align global and local features in multi-level jointly. Yi et al\cite{Yi2020} represent point cloud as voxel and address the LiDAR sampling domain gap for car and pedestrain by converting the problem into surface completion task. Zhao et al\cite{Zhao2020} proposed used ePointDA to to bridge the domain shift at the pixel-level by explicitly rendering dropout noise for synthetic LiDAR and at the feature-level by spatially aligning the features between different domains. Langer et al \cite{Langer2020} fuse sequential labeled lidar scans into a dense mesh and create semi-synthetic data to perform training. Jaritz  et al \cite{Jaritz2020} explore how to learn from multi-modality and propose cross-modal UDA (xMUDA) to adapt 2D images semantic information to  and 3D point clouds.

%% file: sections/our_approach.tex
\input{figures/data_flow.tex}
\subsection{Input Representation}
This paper focuses on the domain adaptation for full-scene semantic segmentation from one real-world Lidar scan dataset to another real-world Lidar scan dataset. Our model used a projection-based model as the segmentor backbone. The input of a projection-based model is the projected LiDAR scan refer to Eq.\ref{eqn:spherical_projection}, where  $r$ is the range, $(x,y,z)$ are the coordinates, $(w,h)$ are width and height of the image, $f$ is the angle of the field of view of Lidar, and $f_{up}$ is the up angle of the field of view. After the projection, we can get a range image and a point index image. Furthermore we also get a 3D coordinate map of the point cloud. Many projection-based models \cite{Wu2018seg,Milioto2019,Cortinhal2020} use the range image, reflectivity map, and coordinate map as input. But the point cloud from the different platforms has different coordinate systems, which is not good for domain adaptation. Meanwhile, the normal map have fixed range for all LiDARs. Besides, according to \cite{Gupta2014}, the normal map can help the model perform semantic segmentation for depth image. Therefore, we use a normal map accompanying with the range image, reflectivity map, as input.
\input{equations/spherical_projection.tex}
\input{figures/inpaint4.tex}
In our experiment, we notice that a lot of stripe pattern appears on projected images and labels (see Fig.\ref{fig:inpaint}(a)), which affects the domain adaptation process. To reduce the negative effect, we pre-process the data (see Fig.\ref{fig:inpaint}(b)). We utilize Closing morphological and subtract operations to locate stripe pattern on mask image. Then we inpaint the reflectivity and range image using the Navier-Stokes inpainting algorithm \cite{Bertalmio2001}, and fill the label using the holes’ nearest neighbors. And the normal map is computed from the inpainted range image.
\subsection{Network Structure}
Generally, we expect a model that can complete a task for data from similar domains. However, feature difference between two similar domains causes a model, which learns from one domain (called source domain \(S\)), can not perform well on another domain (called target domain \(T\)). Therefore, we expect a method that can adapt a model from one domain to another domain. If the target domain does not provide ground truth, the problem is called unsupervised domain adaptation. In this paper, the task is full-scene semantic segmentation for Lidar scan. In this problem, we use \(X_S\) denotes source data, \(Y_S\) denotes source labels, and \(X_T\) denotes target data, but target labels are not accessible.

Based on the intuition that two similar domains should contain shared information across the two domains and private information to each domain. And the adaptable information should be contained in shared information of two domains. Therefore, we designed an end-to end trainable model that splits input data into domain shared and private features. The model then utilizes the extracted shared features to perform semantic segmentation.
The model contains two extractors: a shared feature extractor \(f_P\)  and a private feature extractor \(f_D\) see Fig.\ref{fig:data_flow}. To induce the two extractors to produce such split information, we add a loss function that encourages the independence of these parts, and connect the private feature extractor to a classifier \(f_C\) to differentiate the data from two domains. Besides, we feed the output of the domain shared extractor to a segmentor to complete the same task (predict semantic labels \(\hat{Y}\)).

To ensure that the private features are still useful and to further reduce the domain gap, we introduce the CycleGAN mechanism \cite{Zhu2017} to induce the models to learn two mappings between two domains. The domain private and shared features are fed into domain converters to convert the data from one domain to another domain: (\(f_{S\to T}\) converts source data into target domain, \(f_{T\to S}\) converts target data into source domain). The conversion is learning through an adversarial learning procedure. Therefore, the converted data are separately fed into domain discriminators (\(D_{T}\) and \(D_{S}\)).
Meanwhile, we add Gated-SCNN \cite{Takikawa2019} on the side of the segmentor to extract boundary maps \(B\)  while learning to predict semantic segmentation. The boundary maps are the predicted boundary of semantic labels. We utilize the output boundaries to penalize the label predictions from the target domain. To further penalize the label output, we add a boundaries discriminator \(D_{B}\), and a labels discriminator \(D_{Y}\) to penalize the output label and boundary. 
\subsection{Multi-task Learning}
\label{marker}
The domain adaptation procedure is essentially a multi-task learning procedure. The tasks include domain private feature classification, boundaries extraction, semantic segmentation, domain mutual conversion, similarity measurement of domain shared features and divergence measurement of the domain private features. The complete loss function can be written as follows:
\input{equations/all_loss.tex}
where \(L_{P}\), \(L_{B}\), \(L_{Seg}\), \(L_{M}\), \(L_{C}\) and \(L_{D}\) correspond to the loss of domain private feature classification, boundaries extraction, semantic segmentation, domain mutual conversion, domain similarity and domain difference and \(\lambda_P\), \(\lambda_B\), \(\lambda_{Seg}\), \(\lambda_M\) \(\lambda_C\) and \(\lambda_D\)are hyperparameters that control the weighting between losses. 

The domain private feature classification task is a binary classification problem, which uses standard binary cross-entropy loss Eq. (\ref{eqn:bce_loss}). 
\input{equations/bce_loss.tex}
Therefore, the classification loss is \ref{eqn:classification_loss}, where \(\delta(X) = \{1:x\in X_S;0:x\in X_T\}\).
\input{equations/classification_loss.tex}

For the boundaries extraction task, we have access to labels of source data, which allows us to get the boundaries of source data \(B_S\). We then use standard binary cross-entropy (BCE) loss on predicted boundary maps \(\hat{B}_s\) of source data. From experiments, the network inclines to generate blank results if there was no penalty on target data. Therefore, we add a GAN loss to encourage the network to predict boundaries for target data too. We express GAN loss as Eq.(\ref{eqn:gan_loss})
\input{equations/gan_loss.tex}
Then, the complete loss function of boundary extraction task can be written as Eq.(\ref{eqn:boundary_loss}), where \( G_{B}(x)\) equals \(f_C(x)\) ,and \(\lambda_{B_{GAN}}\), \(\lambda_{B_{BCE}}\) are hyper-parameters for balancing the effect between the GAN loss and BCE loss.
\input{equations/boundary_loss.tex}

For the semantic segmentation task, \(L_{Seg}\) consist of two parts: \(L_{Seg}^S\) of source data and \(L_{Seg}^{T}\) of target data. We employ standard cross-entropy (CE) loss with dual boundary regularizer \cite{Takikawa2019} and Lov\'asz-Softmax loss on predicted labels of source data. 
\input{equations/seg_s_loss2.tex}
Where $G_{Seg}(X)=f_{seg}(f_{C}(X)))$.

For target data, we employ GAN loss to learn segmentation \cite{Luc2016}. Besides, because learning the boundary map is easier than learning semantic segmentation. Therefore, we used the boundary prediction \(\hat{B}_T\) to penalize the label predictions \(\hat{Y}_T\) of the target data. We add a Laplacian layer to extract the boundary of the predicted labels and use the L1 loss to measure the difference between the boundray prediction and the predicted label boundary. In the end, the segmentation loss of source data is Eq.(\ref{eqn:seg_t_loss})
\input{equations/seg_t_loss.tex}

In order to further eliminate the effect of domain difference, we introduce the CycleGAN mechanism into our model, which leads to the fourth task: domain mutual conversion task. We expect that through learning the domain mutual conversion, the model can find the interior relationship between two domains. The mutual conversion task requires two mapping functions: $G_{S \to T}$ maps data from the source domain to target domain, $G_{T \to S}$ maps data from target domain to source domain. The two mappings function can be expressed as \ref{eqn:conversion_fun}. 
\input{equations/conversion_fun.tex}
Based on the two mapping functions, we can define the domain mutual conversion loss as follows:
\input{equations/mutual_loss.tex}
Where $L_{inv}$ and $L_{cyc}$ represent domain invariance loss and cycle consistency loss.

The domain invariance means that the data domain will not be changed if it passes through its domain convertor. For example, we will get data in source domain \(X_{S(S)}\) after data from source domain \(X_S\) pass the mapping function $G_{T \to S}$. This invariance character of the mapping function can be learned through the following function:
\input{equations/invariance_loss.tex}

On the other end, cycle consistency means that after data passes two different mapping functions, its domain should be in its original domain. For example, source domain data \(X_S\) first passes the mapping function $G_{S \to T}$. The converted results \(X_{T(S)}\) passes the mapping function $G_{T \to S}$. We will finally get data \(X_{S(T(S))}\), which should be in the source domain. The cycle consistency loss can be defined as:
\input{equations/cyc_loss.tex}

We meausure the similarity of the shared features between the original data and converted data using \(L1\) loss:
\input{equations/sim_loss}

To measure the divergence of the shared features and private features, we define the loss via as soft subspace orthogonality constraint between the private and shared features \cite{Bousmalis2016}:
\input{equations/diff_loss}
Where \(\parallel \bullet \parallel _F\) is the squared Frobenius norm. \(\mathbf{H^{s}_{p}}\) and are \(\mathbf{H^{t}_{p}}\) are the matrices whose row are the private features from the source domain and target domain respectively. And \(\mathbf{H^{s}_{c}}\) and are \(\mathbf{H^{t}_{c}}\)  are the matrices whose row shared features from the two different domain. 

%% file: figures/data_flow.tex
\definecolor{header}{HTML}{0540F2}
\definecolor{shared}{HTML}{3D6AF2}
\definecolor{private}{HTML}{02732A}
\definecolor{source}{HTML}{FFBE28}
\definecolor{target}{HTML}{F24130}

\begin{figure*}[h]
    \centering
    \includegraphics[width=\textwidth]{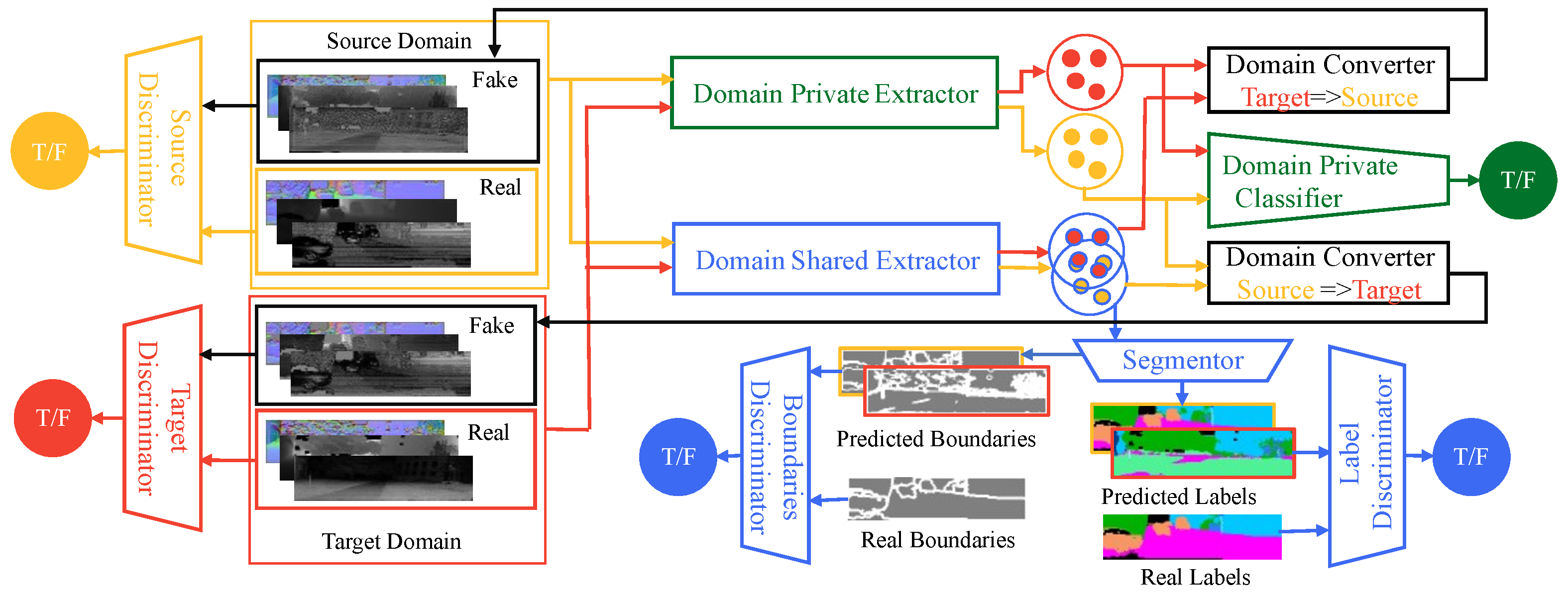}
    \caption{Information Flow Graph: The data (\textcolor{source}{source} and \textcolor{target}{target}) are fed into two branches: one branch is composed of a \textcolor{private}{domain private extractor \(f_P\)} and a \textcolor{private}{domain private classifier \(f_D\)} which can differentiate the input from the two domains, another branch is a \textcolor{shared}{domain shared extractor \(f_C\)}  which extracts the common feature between the two domains including semantic information \(Y\) and boundries information \(B\). A \textcolor{shared}{segmentor \(f_{Seg}\)}  predicts labels \(\hat{Y}\) and boundries information \(\hat{B}\) based on the features from domain shared extractor. The predicted boundaries are sent to a \textcolor{shared}{boundaries discriminator \(D_{B}\)}, and the predicted labels are sent to a \textcolor{shared}{labels discriminator \(D_{Y}\)}. Next, the domain private features and domain shared features are fed into domain converters (\textcolor{target}{\(f_{S\to T}\)} converts source data into target domain, \textcolor{source}{\(f_{T\to S}\)} converts target data into source domain). The coversion are learning through an adversarial learning procedure. Therefore, the coverted data are seperately fed into domain discriminators (\textcolor{target}{\(D_{T}\)} and \textcolor{source}{\(D_{S}\)}). Along with this, the converted data are also fed back to the model to repeat the above procedures. 
    }
    \label{fig:data_flow}
\end{figure*}

%% file: equations/spherical_projection.tex
\begin{equation}
  \label{eqn:spherical_projection}
    \begin{pmatrix}
      u \\ v
    \end{pmatrix}=
    \begin{pmatrix}
      \frac{1}{2}[1-arctan(y,x)\pi ^{-1}]w \\
      [1-(arcsin(zr^{-1})+f_{up})f^{-1}]h
    \end{pmatrix}
  \end{equation}

%% file: figures/inpaint4.tex
\begin{figure}[h]
    \centering
    \includegraphics[width=0.45\textwidth]{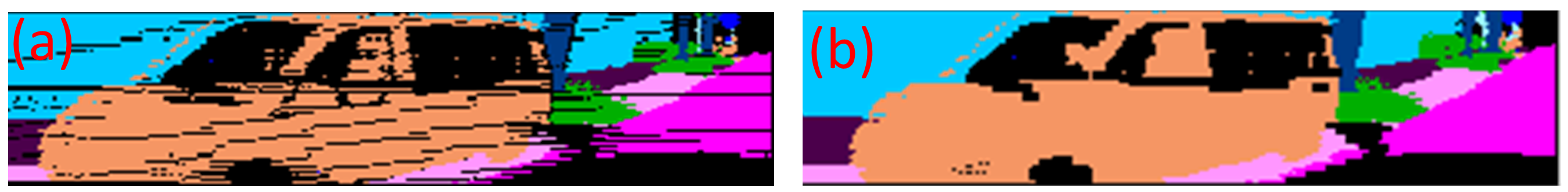}
    \caption{(a) the original label; b) the inpainted label; }
    \label{fig:inpaint}
\end{figure}

%% file: equations/all_loss.tex
\begin{equation}
    \label{eqn:all_loss}
        L = \lambda_PL_{P} + \lambda_BL_{B} + \lambda_{Seg}L_{Seg} + \lambda_ML_{M} + \lambda_CL{C} + \lambda_DL_{D}
\end{equation}

%% file: equations/bce_loss.tex
\begin{equation}
  \label{eqn:bce_loss}
    L_{BCE}(Y,\hat{Y}) = E_{y\sim Y}[y\log{(\hat{y})}+(1-y)log(1-\hat{y})]
\end{equation}

%% file: equations/classification_loss.tex
\begin{equation}
    \label{eqn:classification_loss}
    L_{P}  = L_{BCE}(\delta(X),f_D(f_P(X))
\end{equation}

%% file: equations/gan_loss.tex
\begin{equation}
  \label{eqn:gan_loss}
  \begin{split}
    L_{GAN}(G,D_Y,X,Y) & = E_{y\sim Y}[\log{D_Y(y)}] \\
                       & + E_{x\sim X}[\log(1-D_Y(G(x)))]  
  \end{split}
\end{equation}

%% file: equations/boundary_loss.tex
\begin{equation}
    \label{eqn:boundary_loss}
    \begin{split}
        L_{bd} & = \lambda_{B_{BCE}}L_{BCE}(B_S,\hat{B}_S) \\
        & +  \lambda_{B_{GAN}}L_{GAN}(G_{B},D_{B},X_{T},B_{T})
    \end{split}
\end{equation}

%% file: equations/seg_s_loss2.tex
\begin{equation}
  \label{eqn:seg_s_loss2}
  \begin{split}
    L_{Seg}^{S}= &\lambda_{SS1}L_{CE}(Y_S,\hat{Y}_S) + \lambda_{SS2}L_{dual}(Y_S,\hat{Y}_S,\hat{B}_S) \\
                 & + \lambda_{SS3}L_{Loasz}(Y_S,\hat{Y}_S,\hat{B}_S)
  \end{split}
\end{equation}

%% file: equations/seg_t_loss.tex
\begin{equation}
    \label{eqn:seg_t_loss}
    \begin{split}
        &L_{Seg}^{T}  = \lambda_{ST1}L_{GAN}(G_{Seg},D_{Seg},X_T,Y_S) \\
        &+ \lambda_{ST2}E_{x_t\sim X_T}^{b_t\sim \hat{B}_T}[\parallel Laplacian(G_{Seg}(x_t))-\hat{b}_t\parallel_1]
    \end{split}
\end{equation}

%% file: equations/conversion_fun.tex
\begin{equation}
    \label{eqn:conversion_fun}
    \begin{matrix}
        G_{T \to S}(X)=f_{T \to S}(f_P(f_H(x_t)),f_C(X))\\
        G_{S \to T}(X)=f_{S \to T}(f_P(f_H(x_s)),f_C(X))
       \end{matrix}
\end{equation}

%% file: equations/mutual_loss.tex
\begin{equation}
    \label{eqn:mutual_loss}
        L = \lambda_{M_{inv}}L_{inv} + \lambda_{M_{cyc}}L_{cyc}
\end{equation}

%% file: equations/invariance_loss.tex
\begin{equation}
    \begin{split}
        &L_{inv}(G_{S\to T},G_{T\to S},X_S,X_T)  = \\
        &  E_{x_s \sim X_S}[\parallel \hat{x}_{s(s)}-x_s \parallel _1] \\
        & +  E_{x_t \sim X_T}[\parallel  \hat{x}_{t(t)}-x_t\parallel _1] \\
        & + L_{GAN}(G_{S\to T},D_{T},\hat{X}_{S(S)},X_T) \\
        & + L_{GAN}(G_{T\to S},D_{S},\hat{X}_{T(T)},X_T)
    \end{split}
\end{equation}

%% file: equations/cyc_loss.tex
\begin{equation}
    \begin{split}
        &L_{cyc}(G_{S\to T},G_{T\to S},X_S,X_T)  = \\
        &  E_{x_s \sim X_S}[\parallel \hat{x}_{s(t(s))}-x_s \parallel _1] \\
        & +  E_{x_t \sim X_T}[\parallel  \hat{x}_{t(s(t))}-x_t\parallel _1] \\
        & + E_{x_s \sim X_S}[\parallel G_{Seg}(\hat{x}_{s(t(s))})-G_{Seg}(x_s) \parallel _1] \\
        & +  E_{x_t \sim X_T}[\parallel   G_{Seg}(\hat{x}_{t(s(t))})- G_{Seg}(x_t)\parallel _1] 
    \end{split}
\end{equation}

%% file: equations/sim_loss.tex
\begin{equation}
    \begin{split}
        &L_{C}(f_{C},G_{S\to T},G_{T\to S},X_S,X_T)  = \\
        &  E_{x_s \sim X_S}[\parallel f_{C}(\hat{x}_{t(s)})-f_{C}(x_s) \parallel _1] \\
        & +  E_{x_t \sim X_T}[\parallel  f_{C}(\hat{x}_{s(t)})-f_{C}(x_t)\parallel _1] 
    \end{split}
\end{equation}

%% file: equations/diff_loss.tex
\begin{equation}
    \begin{split}
        L_{D}  = \parallel\mathbf{H^{s}_{c}}^{\top} \mathbf{H^{s}_{p}}\parallel _F +  \parallel\mathbf{H^{t}_{c}}^{\top} \mathbf{H^{t}_{p}}\parallel _F 
    \end{split}
\end{equation}

%% file: sections/experimental_evaluation.tex
\input{./tables/evaluation_table.tex}
\subsection{Dataset}
Recent released datasets \cite{Geyer2020,Che2019,Behley2019,Pan2020} of LiDAR scan for autonomous driving are most collected on vehicles in on-road environments. However, for autonomous robotics, they may have different sensor setup and pass not only on-road environment. These differences will lead to different domain gaps between current existing datasets. Therefore, to verify our method's effectiveness and provide data for research in autonomous robotics in the future. We introduce our dataset SemanticUSL.

SemanticUSL was collected on a Clearpath Warthog robotics with an Ouster OS1-64 LiDAR. The data collection location includes the campus site and off-road research facility of Texas A\&M University. The data include the traffic-road scene, walk-road scene, and off-road scene \ref{fig:dataset_environment}. Our dataset has 16578 unlabeled scans for domain adaptation training and 1200 labeled scans for evaluation. The data uses the same format and ontology as SemanticKITTI \cite{Behley2019}; therefore, it can be easily used for domain adaptation research between SemanticKITTI and SemanticPOSS.  

We evaluated our algorithm on three datasets:  SemanticKITTI dataset \cite{Behley2019}, SemanticPOSS \cite{Pan2020}, and our dataset(SemanticUSL). The information on the other two datasets is as followed:
\begin{itemize}
\item SemanticKITTI is labeled from the KITTI dataset collected around Karlsruhe's mid-size city, rural areas, and highways. The data was collected using a Volkswagen Passat B6 with a  Velodyne HDL-64E.  The Semantic KITTI dataset has 23201 labeled scans and 28 classes.
\item SemanticPOSS was collected at the Peking University campus and contained many dynamic and complex scenes, which are different from SemanticKITTI. The platform is a JEEP with a Pandora 48 channel LiDAR. The dataset has 2988 frames and 14 classes.
\end{itemize}
In summary, the three datasets were collected on different platforms and different sites. These differences cause the divergence of point cloud distributions, noise, and reflectivity, etc.  
\input{figures/dataset_environment.tex}

\subsection{Evaluation Metrics.}
To evaluate performance of our model, we use the widely used intersection-over-union (IoU) metric, mIoU, over all classes \cite{Everingham2014}, given by
\input{equations/mIoU.tex}
where \(TP_c\),\(FP_c\) and \(FN_c\) represent the number of true positive, false positive and false negative predictions for class \(c\) and \(C\) is the number of classes.

\subsection{Implementation Details}
In our experiments, all our networks are implemented using PyTorch\cite{NEURIPS2019_bdbca288}. Training is done on a lambda workstation with two NVIDIA Titan RTX. The LiDAR scans are projected as \(64\times 2048\) resolution range maps. While training, the maps were randomly cropped and resized into a size of \(64\times 512\), with a batch size of 18. The major task model is a semantic segmentation model composed of the shared features extractor and segmentor. In our implementation, we use the encoder of SalsaNext\cite{Cortinhal2020} as the shared feature extractor and its decoder as the segmentor. We add GSCNN along with the second, third, and fourth layers of the segmentor and fuse the boundary output through atrous spatial pyramid pooling module \cite{Chen2018}. The private feature extractor is a copy of the shared feature extractor. And a 2D adaptive average pool layer followed by a fully connected linear layer composes the domain private classifier. The converter uses an atrous spatial pyramid pooling module to fuse the shared and private features and use the decoder of SalsNext to finish the reconstruction. For the discriminators, we use the same PatchGANs in CycleGAN\cite{Zhu2017}. Besides, we also change all the batch normalization layer into instance normalization layer \cite{Ulyanov2016,Zhu2017} for better conversion results. We train the whole model from scratch with a learning rate of 0.001 and an SGD optimizer. The training follows the same style as CycleGAN see Fig.\ref{fig:data_flow}. We first feed real data from the source and target domain into the model and get two predicted labels, two boundary maps, and two fake data. Then, we feed the two fake data into the model and get another six outputs. 

\subsection{Quantitive Evaluation}
In this section, we study our approach's domain adaptation ability among the SemanticKITTI, SemanticPOSS, and SemanticUSL datasets by treating one as the source domain and another as the target domain. We trained original SalsaNext on source data to provide a reference. Besides, we also compared our method with the pixel-level adapted CyCADA method \cite{Hung2018}, because both approaches are using the CycleGAN mechanism. And we also are interested in how well the CyCADA can adapt projected 3D data. We train a SalsaNext model on the adapted data by CyCADA. We reported the results in Table \ref{tbl:eval_res}.

The pixel-level CyCADA methods do not work well on the projected 3D points cloud. See Table.\ref{tbl:eval_res}(third column with "cyckitti/usl/poss"), most of the model can keep the same performance trained with the adapted data but can't generalize to the target domain. In the SemanticPOSS $\rightarrow$ SemanticUSL case, the model can't keep the same performance if trained with the adapted data. This adaptation method tries to visually bring close two domains in global features like reflectivity and norm distribution. However, there's no guidance for local features adaptation. Another reason could be  SemanticPOSS doesn't have enough data to complete CycleGAN conversion. The performance degrades a lot on the transformation with SemanticPOSS. 

On the other hand, see Table.\ref{tbl:eval_res}(third column with "ours" and "ourskitti/usl/poss"), the results shows that our method has better adaptation results. Firstly, after adaptation, the performance of our model on the source domain decreased slightly. Meanwhile, the model recovered over 60\% performance on the target domain. For example, in the case of adapting POSS to USL, the model got 52.18\%mIoU on the SemanticPOSS dataset and got 31.72\% on SemanticUSL. The results indicate the amount of label data in the source domain affects the results. In KITTI $\rightarrow$ USL case, the adapted model's performance on the SemanticUSL dataset is even higher than the model trained with USL data only.
We also provide ablation studies about the effect of the CycleGAN mechanism and the boundary penalty on the adaptation results. Without the CycleGAN mechanism, our model got 36.55\% mIoU by adapting from SemanticKITTI to SemanticUSL. Without the CycleGAN, the IoU of Car decreased to  28.76\% because car detection relies on reflectivity \cite{Wu2019}. The CycleGAN mechanism can adapt the reflectivity features between two domains. By disabling the model's boundary-aware part, the model got 37.20\% mIoU by performing domain adaptation from SemanticKITTI to SemanticUSL. Compared with the model's output after disabling the boundary-aware function, the label with boundary-aware parts has better shapes.

%% file: tables/evaluation_table.tex
\begin{savenotes}
\begin{table*}[t]
\footnotesize  
\setlength{\tabcolsep}{4.5pt}

    \vspace{0.3cm}
    \centering
    \caption{Domain Adaptation Experiment results on SemanticKITTI, SemanticPOSS and SemanticUSL}
    \begin{tabular}{c | c | c | c c c c c c c c c c c c | c}
        \hline
        Source &  Target & 
        Method &  person &  rider &  car  & trunk  &  vegetation & traffic-sign & pole &  object &  building & fence  &  bike &  ground &  mIoU\\
        \hline
        \multirow{11}{*}{KITTI}&\multirow{5}{*}{KITTI}&
        kitti & 62.09 & 74.21 & 93.59 & 61.15 & 91.11 & 37.99 & 57.94 & 50.36 & 84.82 & 54.64 & 15.48 & 94.13 & 64.79 \\
        &&
        cycposs & 64.22 & 76.44 & 92.36 & 60.64 & 90.57 & 37.75 & 57.09 & 46.80 & 84.08 & 51.35 & 15.35 & 93.80 & 64.20 \\
        &&
        cycusl & 58.42 &  69.05 &  92.31 &  56.33 &  90.53 &  37.23 &  56.09 &  44.70 &  82.04 &  47.51 &  13.63 &  93.66 & 61.79  \\
        &&
        oursposs & 47.46 & 68.52 & 94.06 & 73.90 & 47.62 & 37.33 & 59.15 & 58.24 & 88.45 & 27.75 & 29.41 & 56.11 & 57.33\\
        &&
        oursusl & 46.04 & 68.86 & 94.95 & 69.91 & 81.49 & 38.60 & 63.65 & 50.05 & 88.07 & 22.20 & 37.74 & 91.19 & 62.73 \\
        \cline{2-16}
        &\multirow{3}{*}{POSS}&
        source  & 22.77 &  1.78 & 35.91 & 16.86 & 39.84 &  7.08 &  9.73 &  0.18 & 57.03 &  1.64 & 18.17 & 41.99 & 21.08 \\
        &&
        cycada  &  0.00 &   0.00 & 0.00 &  1.45 &  0.00 &  0.00 &  0.00 &  0.00 &  0.00 &  0.00 &  0.00 &  0.00 &  0.12 \\
        &&
        ours  & 31.39 & 23.98 & 70.78 & 21.43 & 60.68 &  9.59 & 17.48 &  4.97 & 79.53 & 12.57 &  0.78 & 82.41  & 34.63\\
        \cline{2-16}
        &\multirow{3}{*}{USL}&
        source   & 33.90 &  0.00 & 27.45 & 10.68 & 36.89 & 16.20 & 12.72 &  5.68 & 41.61 &  3.55 & 31.60 & 75.95 & 24.69 \\
        &&
        cycada   &  0.38 &  0.00 &  28.70 &  13.83  &  57.11 &  20.70 &  23.83 &  3.78 &  53.14 &  22.30 &  9.24 &  72.36 & 25.45 \\
        &&
        ours     & 33.17 &  0.00 & 67.75 &   38.95  & 85.60 &  49.93  & 43.44  & 8.94  & 72.86  & 44.06  & 23.07 & 93.18 & 46.75\\
        \hline\hline
        \multirow{11}{*}{POSS}&\multirow{5}{*}{POSS}&
        source  &  64.47 & 48.25 & 85.77 & 29.71 & 62.71 & 27.29 & 38.19 & 8.07 & 84.90 & 48.50 & 65.56 & 72.56 & 53.00 \\
        &&
        cyckitti  & 64.80 & 48.05 &  84.52 & 29.16 & 61.81 & 26.42 & 33.99 & 8.44 &  84.15 & 47.45 &  65.52 & 71.87 & 52.18 \\
        &&
        cycusl & 32.82 & 17.73 & 71.88 & 16.76 & 53.41 & 16.89 & 16.96 & 1.03 & 71.43 & 24.20 & 49.45 & 67.25 & 36.65 \\
        &&
        ourskitti  & 61.13 & 45.33 & 82.88 & 30.93 & 60.82 & 32.60 & 32.46 &  7.05 & 82.82 & 37.42 & 60.05 & 72.64 & 50.51 \\
        &&
        oursusl & 60.70 &  44.93 &  85.19 &  31.33 &  61.60 &  35.12 &  35.33 &  9.83 &  84.30 &  41.21 &  65.10 &  71.62 & 52.19 \\
        \cline{2-16}
        &\multirow{3}{*}{KITTI}&
        source &   5.20 &  0.50 & 22.57 &  0.54 & 44.00 &  1.90 & 12.83 & 0.08 & 43.09 &  0.70 &  0.40 &  5.62 & 11.45\\
        &&
        cycada & 0.28 &   0.68 &   4.67 &  0.32 & 23.75 &  0.75 &  5.01 &  0.49 &  12.29 &  0.83 &  0.06 &  6.94 &  4.67 \\
        &&
        ours & 23.64 & 24.86 & 73.31 & 23.67 & 72.38 &  4.17 & 31.28 &  2.48 & 59.41 &  0.36 &  0.53 & 68.68 & 32.06 \\
        \cline{2-16}
        &\multirow{3}{*}{USL}&
        source   &   2.45 &  0.00 & 16.15 &  1.21 & 27.94 &  1.34 &  4.52 & 0.62 & 44.37 &  0.12 &  1.16 &  8.05 &  8.99  \\
        &&
        cycada   & 0.00 & 0.00 &   0.00 &  0.05 &  9.40 &  0.19 &  1.12 & 0.15 & 5.06 &  0.28 &   0.00 &  28.01 & 3.69 \\
        &&
        ours  & 30.38 &   0.00 &  45.73 &  28.69 &  63.08 &  22.29 &  33.92 &  4.12 &  63.70 &   1.89 &   9.42 &  77.49 & 31.73 \\
        \hline\hline
        \multirow{7}{*}{USL\footnote{Because there's only 1200 labeled data in SemanticUSL, therefore we use all of the for training and not show the model evalutation results on the SemanticUSL}}&USL&
        source   &   51.96 &  0.00 &  12.57 &  26.29 &  72.89 &  11.18 &  47.22 &  15.11 &  59.78 &  39.61 &  0.00 &  85.61 & 35.19 \\
        \cline{2-16}
        &\multirow{3}{*}{KITTI}&
        source &    1.11 &  0.01 &  28.77 &  5.61 &  38.75 &  3.93 &  15.43 &  1.80 &  29.77 &  1.94 &  1.24 &  46.45 & 14.57\\
        &&
        cycada & 0.17 &  0.00 &  10.43 &  5.06 &  31.86 &  0.53  &  10.26 &  0.96 &  36.31 &  5.42 &  0.14 &  47.47 & 12.38  \\
        &&
        ours & 14.91 &  0.00 &  66.72 &  28.28 &  67.61 &  12.95 &  30.67 &  1.00 &  57.15 &  18.82 &  3.94 &  75.60 & 31.47\\
        \cline{2-16}
        &\multirow{3}{*}{POSS}&
        source  &  4.67 &  0.00 &  21.66 &  3.04 &  27.96 &  1.61 &  6.03 &  0.09 &  41.67 &  2.55 &  5.93 &  63.08 & 14.86 \\
        &&
        cycposs & 5.28 &  0.00 &  9.93 &  5.30 &  25.66 &  1.95 &  5.93 &  0.01 &  52.01 &  0.75 &  0.36 &  50.58 & 13.15\\
        &&
        ours & 11.50 &  0.00 &  44.41 &  19.33 &  41.39 &  5.97 &  13.27 &  0.00 &  71.37 &  0.99 &  1.53 &  66.08 & 22.99 \\
    \hline
    \end{tabular}
\vspace{5pt}
\begin{tablenotes}
  \item\textbf{Table Notes: }  1). \textbf{KITTI} denotes SemanticKITTI;  2). \textbf{POSS} denotes SemanticPOSS;  3). \textbf{USL} denotes SemanticUSL; 4). \textbf{source} denotes ``model was trained on source data only"; 5). \textbf{cyckitti/cycusl/cycposs} denote ``model was trained with pixel-adapted CyCADA method and tested on source domain"; 6). \textbf{ourskitti/oursusl/oursposs}  denotes ``model was trained with our method and tested on source domain"; 7).  \textbf{cycada} denote ``model was trained with pixel-adapted CyCADA method and tested on target domain"; 8). \textbf{ours} denotes ``model was trained with our method and tested on target domain".
\end{tablenotes}
    \label{tbl:eval_res}
\end{table*}
\end{savenotes}

%% file: figures/dataset_environment.tex
\begin{figure}[H]
  \centering
  \includegraphics[width=0.45\textwidth]{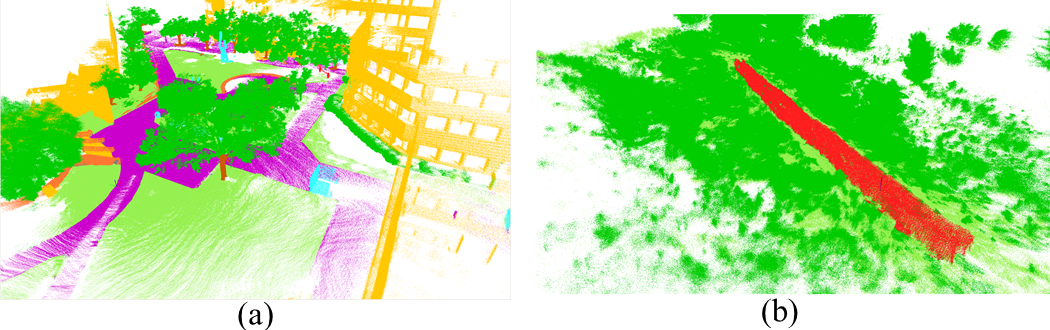}
  \caption{Dataset Collection Environment: (a) is campus environment; (b) is off-road environment}
  \label{fig:dataset_environment}
\end{figure}

%% file: equations/mIoU.tex
\begin{equation}
    mIoU=\frac{1}{C}\sum_{c=1}^{c}\frac{TP_c}{TP_c+FP_c+FN_c}
\end{equation}

%% file: sections/conclusion.tex
In this paper, we propose a boundary-aware domain adaptation approach for semantic segmentation of the lidar point cloud. We design a model that can extract domain shared features and domain private features. We utilize the Gated-SCNN to enable the domain shared feature extractor to keep boundary information in the domain shared features and utilize the learned boundary to refine the segmentation results. We conduct experiments on SemanticKITTI, SemanticPOSS, and SemanticUSL datasets. The results show that our model can keep almost the same performance on the source domain after adaptation and get an 8\%-22\% mIoU performance increase in the target domain.  In future work, we further explore more effective point cloud representation and more efficient architecture to learn the general geometry information.